\documentclass[runningheads]{llncs}
\usepackage[T1]{fontenc}
\usepackage{amsmath,graphicx}
\usepackage{color}
\usepackage{subfigure} 
\usepackage[linesnumbered,ruled,vlined]{algorithm2e}
\usepackage{cite}

\begin{document}
\title{Machine Unlearning Methodology base on Stochastic Teacher Network}
%
%
\author{ Xulong Zhang$^{1}$, Jianzong Wang$^{1}$\thanks{Corresponding author: Jianzong Wang, jzwang@188.com.},  Ning Cheng$^1$, Yifu Sun$^{1,2}$, Chuanyao Zhang$^{1,3}$, Jing Xiao$^1$}
\authorrunning{Zhang et al.}
%
\institute{$^1$\textit{Ping An Technology (Shenzhen) Co., Ltd.}\\$^2$\textit{School of Computer Science, Fudan University}\\$^3$\textit{University of Science and Technology of China }}
\maketitle              
\begin{abstract}

The rise of the phenomenon of the "right to be forgotten" has prompted research on machine unlearning, which grants data owners the right to actively withdraw data that has been used for model training, and requires the elimination of the contribution of that data to the model. A simple method to achieve this is to use the remaining data to retrain the model, but this is not acceptable for other data owners who continue to participate in training. Existing machine unlearning methods have been found to be ineffective in quickly removing knowledge from deep learning models.
This paper proposes using a stochastic network as a teacher to expedite the mitigation of the influence caused by forgotten data on the model. 
We performed experiments on three datasets, and the findings demonstrate that our approach can efficiently mitigate the influence of target data on the model within a single epoch.
This allows for one-time erasure and reconstruction of the model, and the reconstruction model achieves the same performance as the retrained model.

\end{abstract}
\section{Introduction}

Regarding user privacy, recent laws such as the General Data Protection Regulation (GDPR) \cite{voigt2017eu} and the California Consumer Privacy Act (CCPA) \cite{harding2019understanding} have bestowed users with the privilege of exercising their right to be forgotten.
However, deleting user data does not guarantee that the model does not retain any information about that data. This can result in the model making predictions that reveal customer privacy. From a system security perspective, the intentional or unintentional use of low-quality or outdated data during the training phase can exert a detrimental influence on the model.
Previous studies \cite{marchant2022hard,yang2017generative,biggio2012poisoning,rong2022fedrecattack,chen2021machine,caozihao} have shown that the model may be vulnerable to data poisoning attacks from malicious clients. Therefore, there is a need to remove specific data contributions from a trained model to enhance the security and dependability of the model system. However, the relationship between data and specific parameters of the model is not clear in the current field of deep learning \cite{nguyen2022survey,wu2022federated}, making it difficult to modify model parameters directly to effectively mitigate the influence of target data on the model and minimize its impact.

One simple method to forget target data is to delete it from the training dataset and perform model retraining using the remaining data. However, this approach is time-intensive and not feasible for other data owners who continue to participate in training. Machine unlearning \cite{bourtoule2021machine,sekhari2021remember,gupta2021adaptive,kim2022efficient} provides a way to forget target data by effectively mitigating the influence of the forgotten data on the model and reconstructing the model on the remaining data sets. The objective is to achieve the same effectiveness as a retrained model but with a faster and more efficient process.



The main difficulties of machine unlearning come from three aspects\cite{wu2022federated}: the interpretability of the deep learning model \cite{zhang2018visual}, the Randomness of model training process and the Increment of model training process. In \cite {kim2022efficient}, the author proposes to use the contrastive labels to help the model eliminate the impact of target data. In \cite {liu2021federaser}, the author proposes to rebuild the model by  historical parameters. In  \cite {brophy2021machine, baumhauer2022machine} proposed a machine unlearning method for simple machine learning models. But existing machine unlearning methods are difficult to quickly remove knowledge from complex deep learning models.


This paper introduces a novel approach of machine unlearning which can realize fast forgetting of target data and fast reconstruction of model. When a model completely forgets the target data, its performance on the target data should be consistent with that of the model completely untrained by the target data. Therefore, we use a stochastic network that has not been trained by the target data as the teacher network \cite{gou2021knowledge,cho2019efficacy}. By fitting the 
output probability distribution generated by the stochastic network for the target data, we make the trained model to forget the target data. Then, we leverage the original trained model as a mentor network to quickly reconstruct the model on the remaining data set. We have conducted extensive experiments on 3 pulic datasets, 
and the results of our experiments demonstrate that our method can quickly eliminate the impact of target data on the model through only one epoch, achieve the one-time erasure and reconstruction of the model on target data. The reconstructed model has achieved comparable performance to the retrained model.


\section{Related Work}
\subsection{Machine Unlearning}
With the development of artificial intelligence, the model is increasingly dependent on training data, and
there is an increasing number of model parameters and an expanding scale of training data.
Simultaneously, the problem of data ownership is more strict, and the owner of the data  possesses the right to ask for the model to remove the contribution of personal data from its records. From the perspective of security, the system may be attacked by data poisoning, and malicious customers may provide incorrect data to impact the model's performance. The system should possess the capability to mitigate the influence of malicious data on the model through deletion. From the perspective of system availability, The system should allow the customer to recall the wrong data entered unexpectedly. For example, in the recommendation system, the user may enter a certain product by mistake. If the influence of the unexpected data cannot be eliminated, it may lead to frequent recommendation of the wrong product for the user. The objective of machine unlearning is to effectively mitigate the impact of target data on the model.

%

In \cite {bourtoule2021machine}, the author divides the training data into smaller data sets by slicing the data, and reduces the computational overhead of the unlearning phase through the restriction of data point impact on the training process. In  \cite {aldaghri2021coded}, the author proposed to randomly merge the training sample data into a smaller number of samples to train weak learners, which used random projection to model nonlinear relationships, and proposed to use random linear code to speed up the machine unlearning process.
In \cite {liu2021federaser}, the author proposes a machine unlearning method for distributed computing, which helps to accelerate the effect of deleting client data on the global model by saving the historical update data of different clients.
In \cite{wang2022federated}, the author proposes a channel pruning method to eliminate information about specific categories in the model. By quantifying the relationship between the activation of different channels and specific categories in the neural network, the author identifies the corresponding relationship between different channels and specific categories. Then, specific categories are forgotten by pruning the parameters of specific channels. In \cite{kim2022efficient}, the author uses contrastive labels to help the model eliminate the effect of the requested forgotten data on the model. The contrastive label is an error label generated automatically utilizing the output of the original model as a basis for the request for forgotten data. In \cite{wu2022federated}, the author effectively mitigates the influence of client data by directly subtracting the parameters of historical updates from it. However, this approach may lead to model skewness, and to address this issue, the author uses knowledge distillation to reconstruct the model.

\subsection{Knowledge Distillation}
The concept of knowledge distillation \cite{hinton2015distilling}, involves extracting the knowledge encapsulated in a larger teacher model and impart it to a smaller student model. 
This is a knowledge transfer process.
It obtains more output distribution information about the teaching network by introducing soft labels.
In \cite{romero2014fitnets}, to facilitate the training of a more compact student network, the author utilizes the feature maps from the intermediate layer of the teacher network to guide the corresponding student network's layer. In \cite{DBLP:conf/iclr/ZagoruykoK17}, this paper proposes a method to accomplish the transfer of knowledge from the teacher to the student through the attachment map of the network middle layer.
In this work, we employ knowledge distillation as a technique to help the original training model achieve fast data information erasure and rapid model reconstruction.

\section{Methodology}
\subsection{Problem Definition}
We use $D={(x_i,y_i)_{i=0}^n}$ to represent the original dataset participating in model training, where $n$ denotes the total number of training samples, and we use $M_d$ to represent the model trained using the original dataset $D$. $D_f$ indicates the target data requested to be forgotten by the model $M_d$. $D_r$ represents the data remaining after the target data $D_f$ is deleted from the original dataset $D$, where $D=D_f \cup D_r$. The objective is to effectively erase the influence caused by the $D_f$ on the model $M_d$ to obtain the unlearning model $M_u$. One approach to achieve complete eradication of the data from the model is utilizing the remaining data $D_r$ to retrain a model $M_r$. The model retrained under using the remaining data $D_r$ ensures that it does not contain any information about the target data $D_f$. However, the retraining cost required is unacceptable. 
We define a ideal machine unlearning process $L(\cdot)$ as follows:
\begin{equation}
	I(M_r) = I(L(M_d,D,D_f)),
\end{equation}
where $I(\cdot) $ represents a model evaluation function.


\begin{figure*}[tp]
	\centering
	\includegraphics[width=1\textwidth]{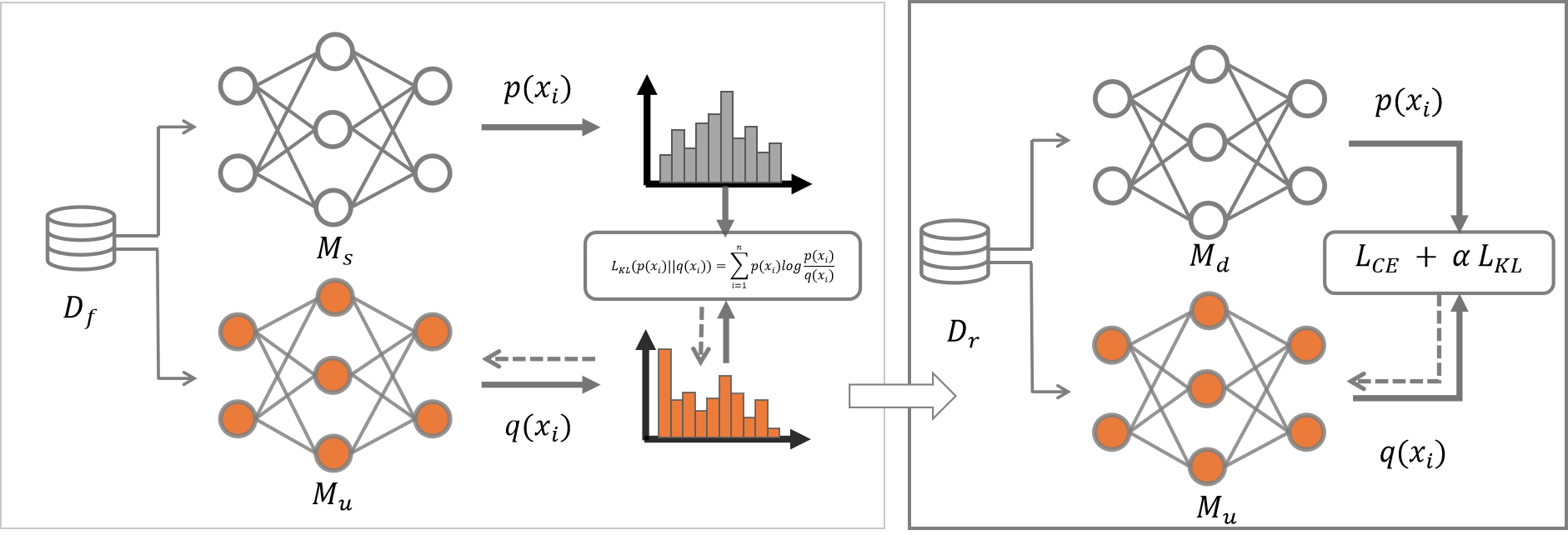}
	\caption{ Our algorithm is divided into two stages. The left represents the stage of   knowledge erasure, and the right represents the stage of  model reconstruction.}
	\label{fig1}
\end{figure*}

\subsection{Our Approach}
We introduce a machine unlearning methodology grounded in stochastic network to help the model eliminate the relevant impact of the target data and can quickly reconstruct the model, so that it can still maintain good performance on the remaining data sets. As shown in Fig.\ref{fig1}. Our algorithm comprises two stages. The left represents the stage of knowledge erasure, and the right represents the stage of model reconstruction.

In the stage of knowledge erasure, the goal is to erasure the relevant impact of forgotten data $D_f$ on the model $M_d$. If the model completely eliminates the impact of the target data $D_f$ requested to be recalled, the classification results of the model $M_u$ for this data $D_f$ should be consistent with the results of random selection. When the classification results are greater or less than the results of random selection, we believe that the model still retains the relevant prior knowledge of $D_f$. We introduce a stochastic initialization model $M_s$, which does not contain any relevant knowledge of $D_f$. We hope $M_u$ to retain the knowledge related to data $D_r$ as much as possible while deleting the relevant knowledge of $D_f$. Since the teacher network is a stochastic initialization network and has not undergone any data training, its classification results of data $D_f$ are random. The results of this ''stupid'' teacher network for target data $D_f$ are exactly what we hope $M_d$ to achieve through machine unlearning. In other words, we hope that machine unlearning algorithms can help $M_d$ forget all relevant knowledge of $D_f$. We use the method of knowledge distillation and use the parameters of $M_d$ to initialize the model $M_u$. $M_s$ will be used as a teacher's network and $M_u$ will be used as a student's network. Through knowledge distillation, we can help the student network $M_u$ forget the relevant knowledge of the target data $D_f$.

We use the parameters of $M_d$ to initialize the model $M_u$, and make the probability distribution of $M_u$ for the dataset $D_f$ consistent with the probability distribution of the stochastic initialization model $M_s$ for the target data $D_f$.

%
As shown on the right of Fig.1, the model $M_s$ and model $M_u$ are composed of convolution neural network with the same structure. We use the parameters of the model $M_d$ trained with complete data to initialization model $M_u$ and random initialization the parameters of model $M_s$. The goal of knowledge erasure is to use $M_s$ as teacher network help model $M_u$ delete the impact of request forgotten data $D_f$ on the model $M_d$. For a sample $x_i\in D_f$, passing $x_i$ through model $M_s$ and $M_u$ respectively to get probability distributions $p(x_i) $ and $q(x_i) $. We use Eq. (\ref{eq:soft}) to obtain the soft label of the model for the samples in the dataset $D_f$, where the temperature parameter $\tau$ used in the softmax function can help to obtain more information about the distribution, $z_i$ represents the probability values of $x_i$ for different categories, and $n$ denotes the total number of categories of datasets $D$.
    
\begin{equation}
    	p(x_i)= \frac{exp(z_i/\tau)} {\sum\limits_{j=0}^{n}exp(z_j/\tau)}  
\label{eq:soft}
 \end{equation}

We use Eq. (\ref{eq:KL}) to calculate the disparity in probability distribution between model $M_u$ and $M_s$ with respect to $D_f$ by Kullback-Leibler Divergence (KLD) and update the parameters of $M_u$ to forget about the knowledge about $D_f$ in the model $M_d$.  

\begin{equation}
	L_{KL}(p(x_i)||q(x_i))= \sum\limits_{i = 1}^{n} p(x_i)  \log \frac{p(x_i)} {q(x_i)}
 \label{eq:KL}
\end{equation}

In the rapid model reconstruction stage, due to the modification of model $M_u$ parameters,  can result in a deterioration of the performance of the model $M_u$ on data $D_r$ while forgetting the relevant knowledge of the target data $D_f$. We use the remaining data set $D_r$ and the original model $M_d$ to help the model $M_u$ rebuild quickly to restore the performance of the model $M_u$. As shown on the right of Fig.1, the original model $M_d$ is regarded as a teacher network, and the model $M_u$ is regarded as a student network. For a sample $x_i\in D_r$, we use Eq. (\ref{eq:KL}) to calculate the KLD of the probability distribution output $p(x_i)$ and $q(x_i)$ by $M_d$ and $M_u$ on dataset $D_r$. At the same time, the Cross Entropy (4) is used as the loss function of $M_u$ on $D_r$. 

 \begin{equation}
 	 L_{CE}=-\sum\limits_{i=0}^{n} y_i log(q(x_i))
 \end{equation}

The total loss is shown in Eq. (\ref{eq:total}), where $\alpha$ is the hyperparameter. We utilize the total loss to aid the model $M_u$ fine tune parameters to quickly restore the performance on $D_r$. The knowledge erasure and model reconstruction are shown in Algorithm 1.
    \begin{equation}
   	L=L_{CE} + \alpha L_{KL}
    \label{eq:total}
   \end{equation}

\begin{algorithm}[hb]
	\label{arg1}
	\caption{ Knowledge Erasure and Model Reconstruction}
	\SetKwInput{KwInput}{Input}                
	\SetKwInput{KwOutput}{Output}              
	
	\DontPrintSemicolon
	
	\KwInput{Request forgotten data$D_f$, Remaining data $D_r$, Original model$M_d$, Stochastic initialization network $M_s$. }
	\KwOutput{Model unlearning relevant knowledge of forgotten data:$M_u$}
	
	\SetKwFunction{Fsever}{Knowledge Erasure}
	\SetKwFunction{Fclient}{Model Reconstruction}
	
	\SetKwProg{Fn}{}{:}{}
	\Fn{\Fsever{$D_f$, $M_s$, $M_d$}}{
		Random initialization model $M_s$ parameters \;
		Using $M_D$ parameters to initialization model $M_u$\;
		\For{ batch $x \in D_f$}{
			
			Compute $L_{KL}$ loss by Eq.(3) \;
			Update $M_u$ parameters with gradient descent \;
		}
		\KwRet $M_u$  \tcp{scratch model}  
	}
	
	\SetKwProg{Fn}{}{:}{}
	\Fn{\Fclient{$D_r$, $M_u$, $M_d$}}{
		\For{batch $x$ $\in D_r$ }{
			Compute $L$ loss by Eq.5\;
			Update model $M_u$ parameters according to $L$ \;
			
		}
		\KwRet $M_u$ \;
	}
	
\end{algorithm}

\section{Experiments and Results}

\subsection{Datasets}
We conduct experiments on three popular benchmarks: CIFAR-10\cite{krizhevsky2009learning},  MNIST\cite{lecun1998mnist} and Fashion-MNIST\cite{xiao2017fashion}.  

The CIFAR-10 dataset comprises 60,000 color images, each with a size of 32$\times$32 pixels. The dataset consists of 10 classes, with each class containing 6,000 images. 

The MNIST and Fashion-MNIST datasets consist of 70,000 28$\times$28 images each, respectively, with 10 classes and 7,000 images per class. The MNIST dataset consists of 60,000 images for training and 10,000 images for testing, while the Fashion-MNIST dataset has the same split of training and testing images.
	
\subsection{Training Details}

Our goal is to verify that our method can effectively erase information from the model related to the target data $D_f$ and quickly reconstruct the model instead of obtaining the highest classification accuracy on a specific dataset. Therefore, we use a simple model architecture, which includes two convolution layers and a full connection layer. Firstly, we train an original model on the complete dataset $D$, and then utilize our algorithm to make the model $Original$ unlearn the relevant information of the requested data $D_f$ to obtain model $Scratch$. Then we use the remaining datasets $D_r$ to help the model $Scratch$ rebuild quickly to restore the model performance. We studied two cases of removing 10\% and 20\% data, and The reconstruction time and forgetting effect are compared with the method of retraining the model.

\subsection{Evaluation Results}

To showcase the efficacy of our approach in quickly erasing relevant information of requested forgotten data, we compared our model with the Original model, the Scratch model, and the Retrained model. In Table \ref{table1}, we present a detailed analysis of the erasure effect of our method when deleting a certain category. The second to tenth rows of Table \ref{table1} represent the classification accuracy of each category in the remaining data, "Remaining data" represents the average classification accuracy of the remaining categories, and "cat" represents the target data requested to be forgotten. We list the accuracy of the original model, the Scratch model, the Retrained model, and the unlearning model obtained by our method in detail in Table \ref{table1}.
\begin{table}[hbt]
	\centering
	\caption{The accuracy of each class in CIFAR-10. The requested forgotten data $D_f$ accounts for 10\% of the total data $D$.}	
			
			
			
 \renewcommand\arraystretch{1.3}
 \begin{tabular}{l|c c c c}
			\hline
			\textbf{Categories}   & \textbf{Original}  & \textbf{Scratch} & \textbf{Retrained} &  \textbf{Our} \\ 
			\hline
			
			Plane      &  68.80\%  & 75.80\%    &  81.30\%    & 79.70\%  \\ 
			Car        & 85.30\%   & 91.50\%     & 84.20\%     & 83.80\%  \\ 
			Bird       & 65.20\%   & 54.80\%     & 56.00\%    & 67.80\%   \\ 
			
			Deer       & 69.00\%   & 64.90\%     &  70.20\%    & 77.80\%  \\ 
			Dog        & 69.00\%   & 45.60\%     &  83.90\%    & 78.10\%  \\
			Frog       & 85.00\%   & 74.50\%     &  86.80\%    & 85.00\%  \\ 
			Horse      & 79.10\%   & 80.40\%     &  76.00\%    & 80.00\%  \\ 
			Ship       & 89.10\%   & 90.10\%     &  85.80\%    & 85.50\%  \\ 
			Truch      & 79.50\%   & 74.00\%     &  80.30\%    & 83.80\% \\ 
			
			\hline
			Remaining data     & 76.67\%   & 72.40\%     & 78.28\%    &  \textbf{80.11\%} \\
			\hline
			\textbf{Cat(Forgotten)}& 49.70\%     & 8.10\%     &  0.00\%   & \textbf{11.40\% }\\  
			\hline
		\end{tabular}
	\label{table1}
\end{table}
The table shows that our method can quickly erase the relevant knowledge of the target data "cat" in the knowledge erasure stage. After the knowledge erasure stage, the classification accuracy of the Original model for the "cat" category decreased from 49.70\% to 8.10\%, while the average classification accuracy of the remaining data decreased from 76.67\% to 72.40\%, and the classification accuracy only decreased by 4.27\%. This indicates that our method can accurately erase specific knowledge in the model. After the model reconstruction stage, our model for the "cat" category was restored to 11.40\%, which is equivalent to the result of random classification for the CIFAR-10 dataset containing 10 categories. Moreover, the average classification accuracy of the remaining dataset increased from 72.40\% to 80.11\%.

Retrained results were acquired by training the model again using the remaining dataset. Since the "cat" class did not participate in the training during the retraining process, all "cat" instances in the testing process were incorrectly predicted by the Retrained model, which we believe is due to the inductive bias of the model itself. The average classification accuracy of the Retrained model in the remaining dataset is 78.28\%. Compared with the Retrained model, our method quickly erases the relevant knowledge of the target data set and achieves better results on the remaining data.

We show the average classification accuracy of our method for the remaining data sets after information erasure and model reconstruction in Table \ref{table2} and \ref{table3}. 
\begin{table*}[h]
	\centering
	\caption{Average test accuracy for remained dataset $D_r$ of all \textit{Dataset}, The requested forgotten data $D_f$ accounts for 10\% of the total data $D$. Blue text indicates training epochs}	
\renewcommand\arraystretch{1.3}
\begin{tabular}{l|lll}
			\hline
			\textbf{Baselines}   & \textbf{CIFAR-10} & \textbf{MNIST}  &  \textbf{FashionMNIST} \\ 
			\hline
			Original             &  76.67\%                            &  99.25\%             & 91.98\%   \\ 
			Scratch              & 72.40\%       					   &  97.48\%             & 89.21\%  \\ 
			Retrained            & 78.28\%  \textcolor{blue}{(20)}     &  99.19\%\textcolor{blue}{(10)}    & 92.88\% \textcolor{blue}{(20)}\\ 
			\hline
			\textbf{Ours}         & \textbf{80.11}\% \textcolor{blue}{(1)}    &  \textbf{99.21}\% \textcolor{blue}{(1)}  &\textbf{94.13}\% \textcolor{blue}{(1)} \\
			\hline 
		\end{tabular}	
	\label{table2}
\end{table*}

\begin{table}[h]
	\centering
	\caption{Average test accuracy for remained dataset $D_r$ of all \textit{Dataset}, The requested forgotten data $D_f$ accounts for 20\% of the total data $D$. Blue text indicates training epochs}	
	\renewcommand\arraystretch{1.3}
		\begin{tabular}{l|lll }
			\hline
			\textbf{Baselines}   & \textbf{CIFAR-10 }  & \textbf{MNIST}  &  \textbf{FashionMNIST} \\ 
			\hline
			Original             &  75.11\%                         &  99.34\%             & 91.01\%   \\ 
			Scratch              &  60.35\%                         & 91.85\%              & 83.66\%\\ 
			Retrained            &  79.82\%  \textcolor{blue}{(20)} &  99.41\% \textcolor{blue}{(10)}        & 93.26\% \textcolor{blue}{(20)}\\ 
			\hline
			\textbf{Ours}         & \textbf{81.45}\% \textcolor{blue}{(1)}  &  \textbf{99.42}\% \textcolor{blue}{(1)}  &\textbf{93.76}\%\textcolor{blue}{(1)} \\
			\hline 
		\end{tabular}

	\label{table3}
\end{table}

Table \ref{table2} illustrates the outcomes of the request to forget 10\% of the data (deleting one category of data), and Table \ref{table3} displays the outcomes of the request to forget 20\% of the data  (deleting two categories of data ). It can be seen from the Table\ref{table2} and Table\ref{table3} that after the knowledge erasure stage of the original model, the accuracy of classification for the remaining data has decreased. After the model reconstruction, the accuracy of the remaining data have been restored. On three data sets, we request to forget one or two types of data, and our model has achieved better performance than the Retrained model. The blue font represents the epochs of the model training. We can see that our model only needs one epoch to quickly recover performance, and retraining the model requires more epochs. We have studied the time required for retrained model and our method in detail in Fig.2.



\subsection{The Speed of Machine Unlearning}
Training speed is an important factor in machine unlearning. It is essential to quickly erase the knowledge of requested forgotten data and regain the model's performance on the remaining data. In order to showcase the efficacy of our proposed method, we conducted experiments on a dataset with two requested forgotten categories. In Fig.2 (a)-(c), we present the classification accuracy of each category after the knowledge erasure and model reconstruction stages. We trained our model using one epoch of knowledge erasure to erase the information of requested forgotten categories, and after one epoch of model reconstruction, our model quickly recovered the performance on the remaining data. Moreover, the classification accuracy of the requested forgotten data remained at random classification level, indicating that our model could erase the knowledge of requested forgotten data effectively.

In Fig.2 (d)-(f), we compare the training speed of our method with the retrained model. The plots show the changes in the accuracy of the remaining data with the training epochs. Our model outperforms the Retrained model in terms of training speed and accuracy. After one epoch, our model quickly recovered the performance of the Scratch model, whereas retraining the model requires more time. Additionally, our model achieved better results than the Retrained model. Based on these experimental findings, it can be concluded that our method is indeed effective in quickly erasing the knowledge of requested forgotten data and restoring the model's performance on the remaining data.

\begin{figure*}[!h]
	\centering
	\subfigure[CIFAR-10 (20\%)]{\includegraphics[width=6cm]{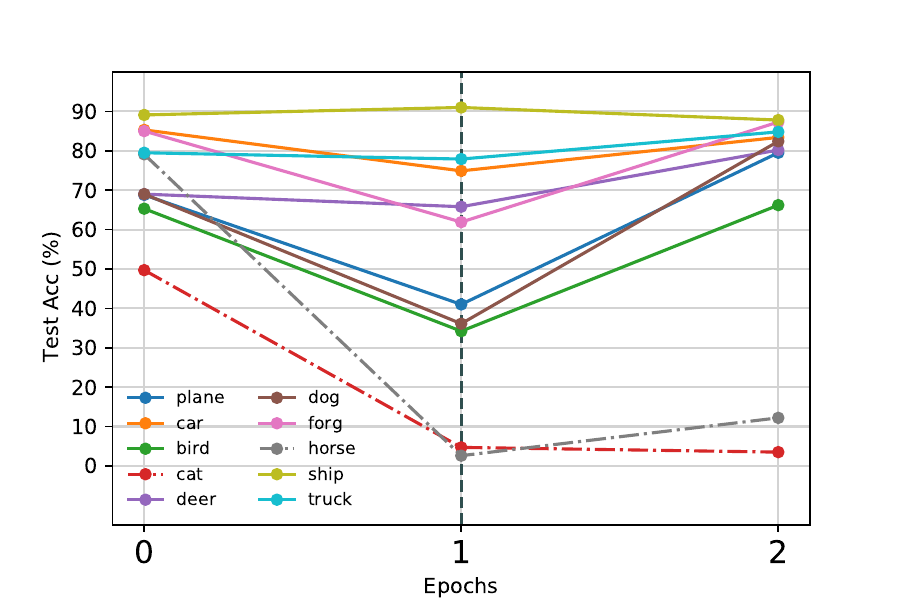}}
	\subfigure[MNIST(20\%)]{\includegraphics[width=6cm]{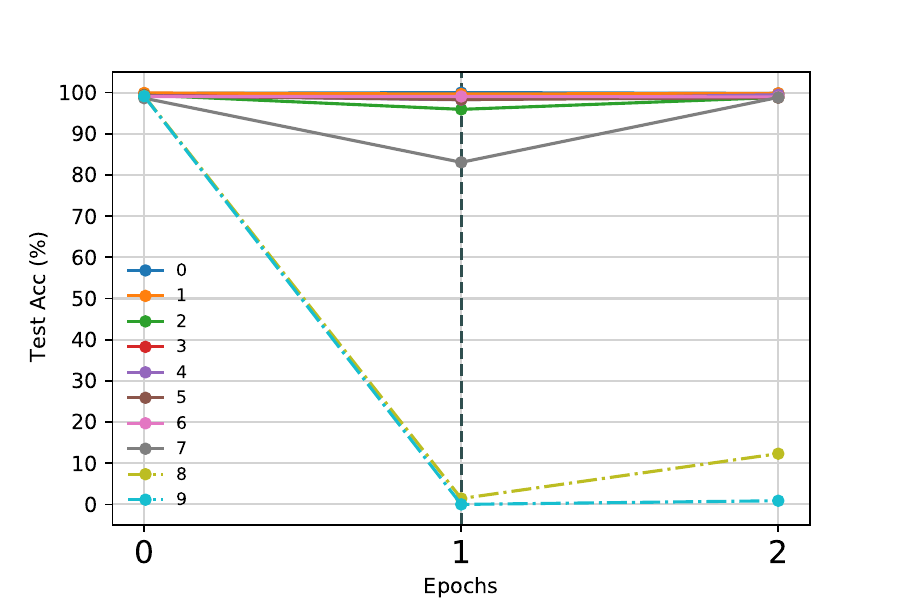}}
	\subfigure[Fashion-MNIST(20\%)]{\includegraphics[width=6cm]{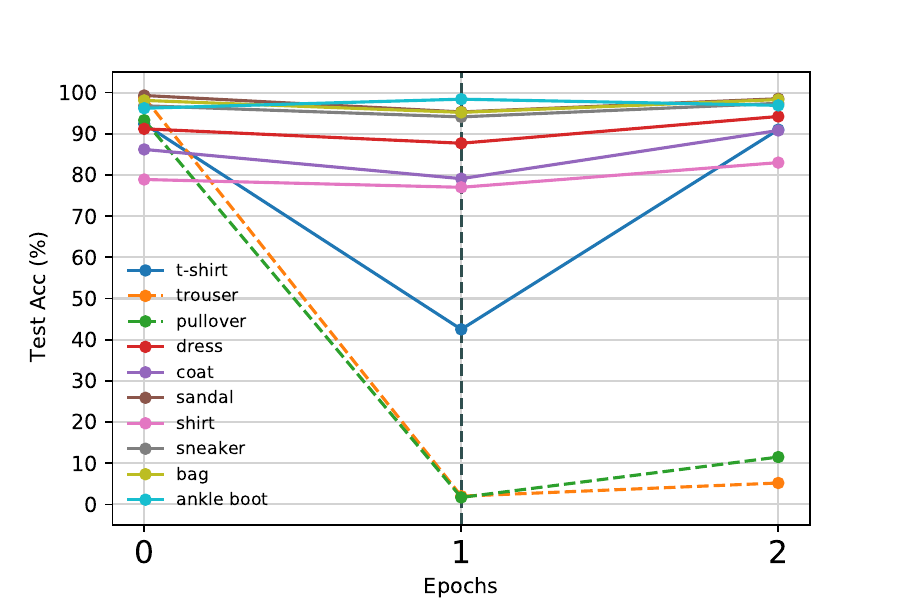}} 
	\subfigure[CIFAR10 (20\%)]{\includegraphics[width=6cm]{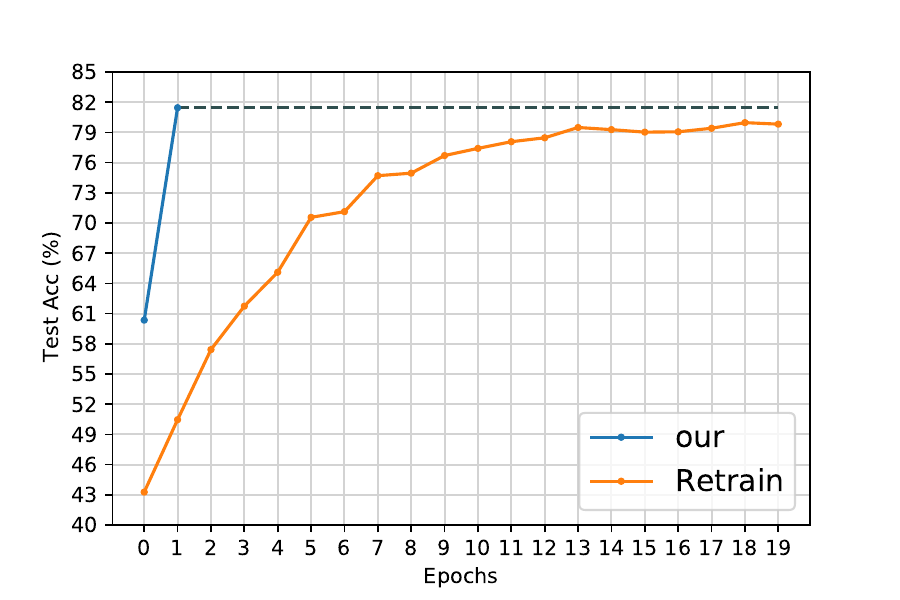}}
	\subfigure[MNIST (20\%)]{\includegraphics[width=6cm]{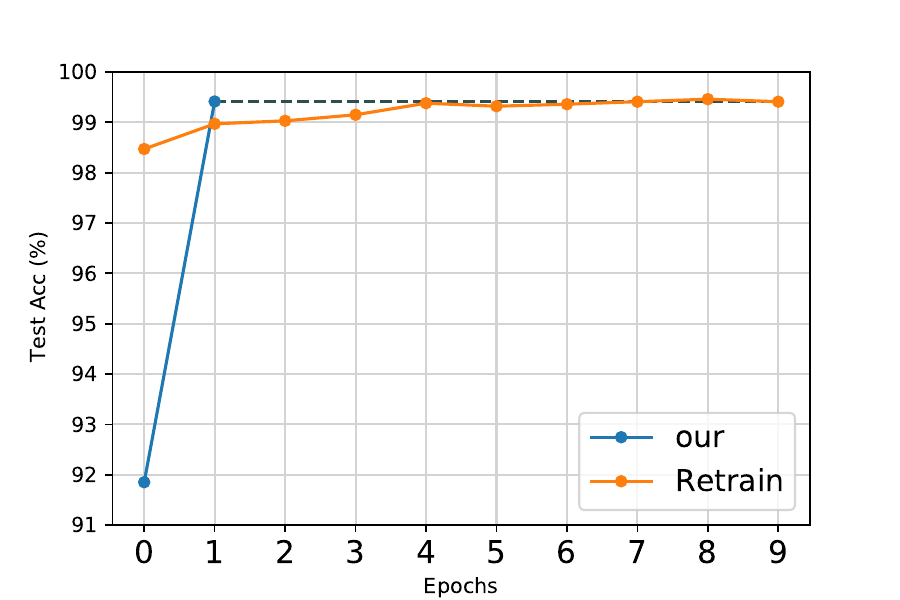}}
	\subfigure[Fashion-MNIST (20\%)]{\includegraphics[width=6cm]{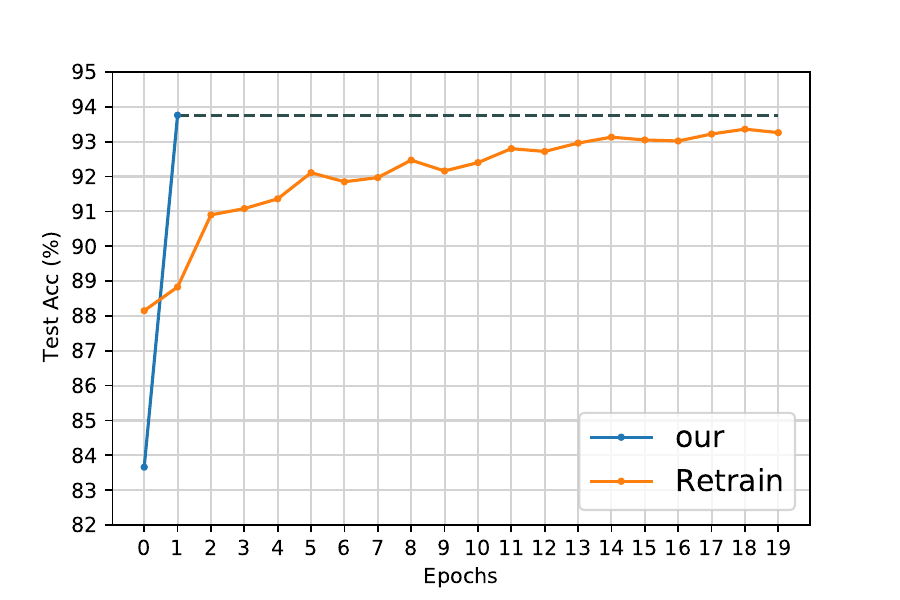}}
	\caption{(a),(b),(c) represent the change of classification accuracy of each category during information erasure and model reconstruction with the training epoch. (d),(e),(f) represent the process of model reconstruction and model retraining.} 
	\label{fig2}
\end{figure*}


\section{Conclusion}
	
This paper presents a novel approach to quickly erase the impact of target data information on the model. Our method is based on an intuitive understanding that when a model completely forgets the target data, its performance on the target data should be consistent with that of the model completely untrained by the target data.
We propose to use a stochastic initialized network as a teacher network to help the original model forget the target data information. Then, we use the remaining data to help model quickly rebuild. We conduct experiments on three benchmarks, and study the accuracy of different categories in the datasets during the knowledge erasure process.
The results validate the efficacy of our method for erasing target data information. Our approach enables rapid mitigation of the influence exerted by the target data on the model through only one epoch and achieve the one-time erasure and reconstruction of the model on target data.

\section{Acknowledgement}
This paper is supported by the Key Research and Development Program of Guangdong Province under grant No.2021B0101400003. Corresponding author is Jianzong Wang from Ping An Technology (Shenzhen) Co., Ltd (jzwang@188.com).

\bibliographystyle{splncs04}
\bibliography{mybib,0-citation-self}

\end{document}